\documentclass[letterpaper]{article} % DO NOT CHANGE THIS
\usepackage[submission]{aaai23}  % DO NOT CHANGE THIS
\usepackage{times}  % DO NOT CHANGE THIS
\usepackage{helvet}  % DO NOT CHANGE THIS
\usepackage{courier}  % DO NOT CHANGE THIS
\usepackage[hyphens]{url}  % DO NOT CHANGE THIS
\usepackage{graphicx} % DO NOT CHANGE THIS
\usepackage{caption} % DO NOT CHANGE THIS AND DO NOT ADD ANY OPTIONS TO IT
\usepackage{algorithm}
\usepackage{algorithmic}
\usepackage{amsmath}
\usepackage{booktabs}
\usepackage{multirow}
\usepackage{amsthm}
\usepackage{amssymb}

\usepackage{newfloat}
\usepackage{listings}
\DeclareCaptionStyle{ruled}{labelfont=normalfont,labelsep=colon,strut=off} % DO NOT CHANGE THIS
\floatstyle{ruled}
\newfloat{listing}{tb}{lst}{}
\floatname{listing}{Listing}
\title{Oracle-guided Contrastive Clustering}

\author{
    Mengdie Wang,
    Liyuan Shang, 
    Suyun Zhao, 
    Yiming Wang, 
    Hong Chen, 
    Cuiping Li,
    Xizhao Wang
}

\begin{document}
\bibliographystyle{apalike}

\maketitle

\begin{abstract}
Deep clustering aims to learn a clustering representation through deep architectures. Most of the existing methods usually conduct clustering with the unique goal of maximizing clustering performance, that ignores the personalized demand of clustering tasks.% and results in unguided clustering solutions. 
However, in real scenarios, oracles may tend to cluster unlabeled data by exploiting distinct criteria, such as distinct semantics (background, color, object, etc.), and then put forward personalized clustering tasks. To achieve task-aware clustering results, in this study, Oracle-guided Contrastive Clustering(OCC) is then proposed to cluster by interactively making pairwise ``same-cluster" queries to oracles with distinctive demands. Specifically, inspired by active learning, some informative instance pairs are queried, and evaluated by oracles whether the pairs are in the same cluster according to their desired orientation. And then these queried same-cluster pairs extend the set of positive instance pairs for contrastive learning, guiding OCC to extract orientation-aware feature representation. Accordingly, the query results, guided by oracles with distinctive demands, may drive the OCC's clustering results in a desired orientation. Theoretically, the clustering risk in an active learning manner is given with a tighter upper bound, that guarantees active queries to oracles do mitigate the clustering risk. Experimentally, extensive results verify that OCC can cluster accurately along the specific orientation and it substantially outperforms the SOTA clustering methods as well. To the best of our knowledge, it is the first deep framework to perform personalized clustering.
\end{abstract}

\section{Introduction}

Clustering, as one of the most fundamental unsupervised learning techniques\cite{Xu2015ACS}, has been successively used in a wide range of applications% in various fields of artificial intelligence
, such as image processing\cite{Cai2011HeterogeneousIF}, gene analysis\cite{Gao2006DiscoveringSO} and text categories\cite{Aggarwal2012ASO}. Recently, by employing highly non-linear latent representations\cite{Min2018ASO}, Deep Clustering (DC) is widely studied and achieves promising clustering results\cite{Xie2016UnsupervisedDE, Caron2018DeepCF,Li2021ContrastiveC,Zhong2021GraphCC,Liu2022DeepGC}. %Although DC has achieved promising clustering performance\cite{}, Some DC methods exploit loss functions related to traditional clustering guiding the deep networks to learn clustering-friendly representations\cite{Xie2016UnsupervisedDE, Caron2018DeepCF}. There are also methods that raise new techniques,  such as contrastive learning\cite{Li2021ContrastiveC, Zhong2021GraphCC} and deep graph clustering\cite{Liu2022DeepGC}. 
Typically, these existing clustering and DC techniques share a common and unique goal, to maximally enhance the clustering performance. However, the personalized demand of clustering tasks is mostly ignored and dismissed.

\begin{figure}[t]
\centering
\includegraphics[width=0.9\columnwidth]{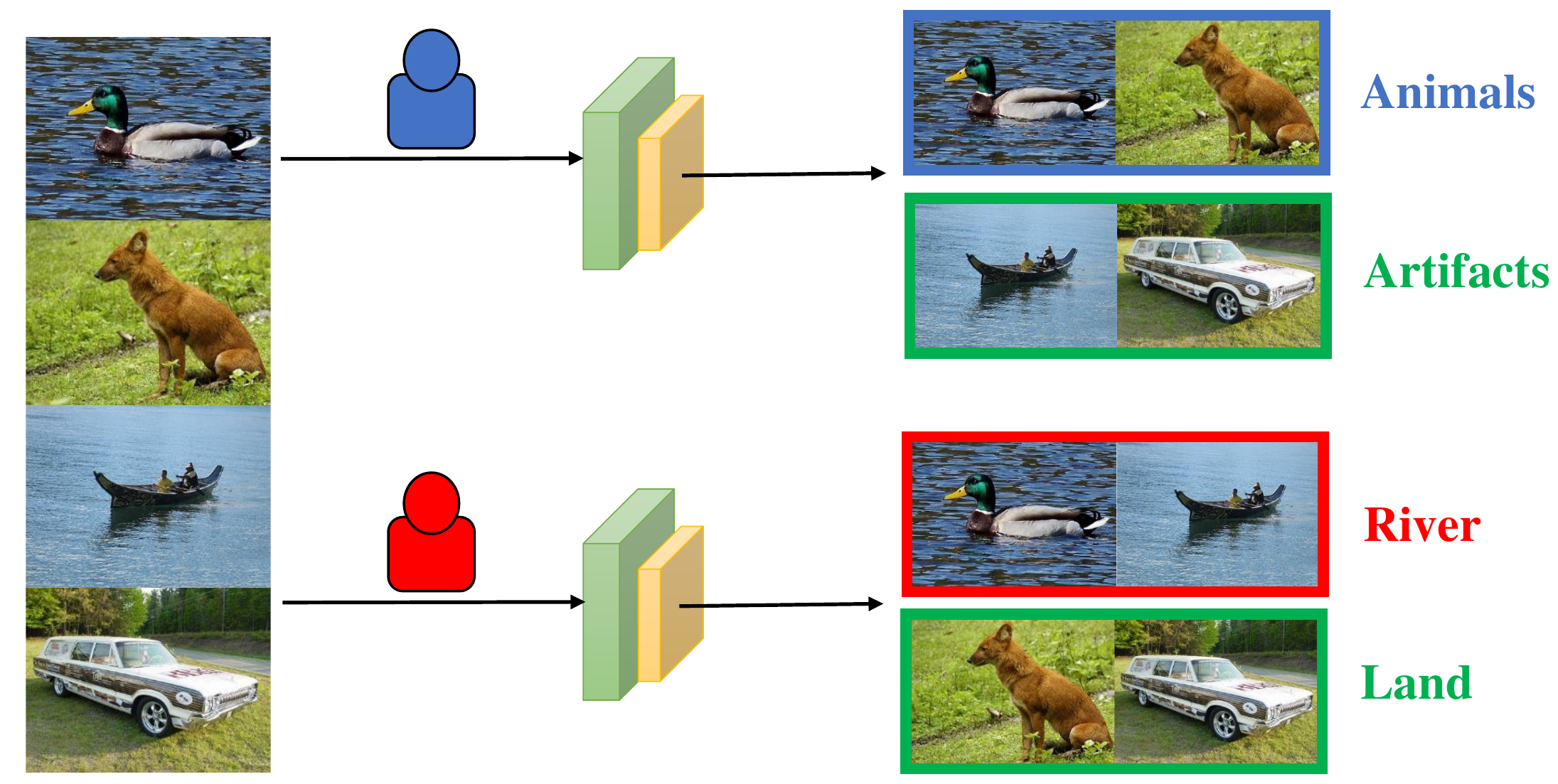}
\caption{The diversity of clustering orientation. The two oracles make different judgments on the task of dividing the four images into two clusters.}.
\label{fig1}
\end{figure}

In the real applications, there are more than one available cluster demands. Fig. \ref{fig1} shows an example of the diversity of clustering orientations. Some tasks require clustering in terms of objects, so the oracle indicates that duck and fox belong to the same cluster, categorized as animals. Some other tasks, however, demand distinguishing backgrounds, the oracle then indicates duck and sailboat belong to the same cluster, categorized as river. % The existing clustering methods can only distinguish images based on the primary features, resulting in unguided clustering results. Thus these methods provide invalid clustering results for clustering tasks focusing on inapparent features. It indicates that how to extract features selectively and cluster along a given orientation is still an open problem. 
In such cases with personalized demands, the existing clustering techniques, clustering with default orientation, may decline or even be unworkable without the guide of oracles. Accordingly, it is still a challenging problem to cluster along a desired orientation. 

To solve this problem, in this study, we propose an Oracle-guided Contrastive Clustering(OCC) model to achieve task-aware clustering results in the guide of oracles with distinctive demands. As active learning, effectively and interactively engaging human for improving annotating performance, benefits us for clustering in the desired orientation, it is exploited to actively select informative instance pairs for oracle to provide answers of same-cluster queries in the forms of yes or no. Thus, the distinctive demand of oracles is embedded in clustering with ``queries and answers" in the loop. To catch orientation-aware feature information in deep clustering, the queried same-cluster pairs are regarded as extensions of the positive instance pairs in the process of contrastive learning, prompting the network to learn shared features between same-cluster pairs. Consequently, with the learned orientation-aware features, the clustering solution is achieved along the personalized demands.

%Same-cluster sample pairs imply the criteria for clustering and reflect the clustering intention of oracles.
%Similar features between same-cluster pairs are learned by contrastive learning to adjust the clustering target. Contrastive learning\cite{Ji2019InvariantIC} seeks to learn a feature representation invariant to data augmentations(positive instance pairs). Inspired by this characteristic, the queried same-cluster pairs are regarded as extensions of the positive instance pairs, prompting the network to learn similar features and cluster along the desired orientation.%A contrastive network learns the semantic features of similar samples by narrowing the distance between positive instance pairs while widening the distance between negative pairs.%
%To take advantage of this characteristic, we explore a pairwise query method to select a pair of images and ask oracles whether they are in the same cluster. The queried same-cluster pairs are extensions of positive instances, guiding the contrastive network to learn specific features about the criterion of clustering.%

Specifically, OCC first constructs the initial positive instance pairs via random data augmentations. Then some instance pairs are selected for annotation according to their similarity and intensity of variation. %The oracle judges whether the instance pair belongs to the same cluster. 
Subsequently, the same-cluster sample pairs annotated by the personalized oracle extend the positive instance pairs in the contrastive loss. After that, the feature output by the backbone network is projected into the representation space and the assignment space, wherein OCC learns shared features between the positive instance pairs by active contrastive losses. The features are orientation-aware, and thus personalized clustering is achieved in the desired orientation. Our major contributions can be summarized as follows:

\begin{itemize}
    
    \item We propose a novel model, named OCC, that exploits active learning joint with contrastive learning to catch the desired feature, and to guide the orientation of clustering according to the personalized demand. 
    \item To the best of our knowledge, it is the first work that concerns the diversity of clustering orientation in deep clustering. Unlike the existing clustering methods, the proposed OCC is bi-objective, that is, to maximize the clustering performance and to accurately cluster in a specific orientation. 
    \item By strict theoretical analysis, a tighter upper bound, that can be mitigated by active query with a specific orientation, is given for the clustering risk in an active learning manner. Simultaneously, extensive experiments demonstrate that OCC can learn clustering features in a targeted manner.
   
\end{itemize}

\section{Related Work}

\subsection{Deep Active learing}
Active learning\cite{Ren2022ASO} aims to query the optimal samples in the unlabeled dataset to reduce the cost of labeling as much as possible while still maintaining performance. The most common query strategies are uncertainty-based approaches, which select samples with high information content to decrease labeling costs. For example, ENS-VarR\cite{Beluch2018ThePO} uses Monte-Carlo dropout and deep ensembles to obtain well-behaved uncertainty estimates from deep neural networks. Ranganathan et al. \cite{Ranganathan2017DeepAL} make efforts to integrate an active learning based criterion in the loss function used to train a deep belief network. 
Density-based approaches have also been applied to CNNs. Core-Set\cite{sener2018active} chooses several scattered center points to minimize the max distance between a data point and its nearest center. TOD\cite{Huang2021SemiSupervisedAL} is a task-agnostic approach based on the observation that the samples with higher loss are usually more informative to the model than that with lower loss. The method propose an effective loss estimator Temporal Output Discrepancy to query samples with higher loss.

\subsection{Active Clustering}
Active learning is also widely used in the field of clustering\cite{Basu2008ConstrainedCA}. Dasgupta and Hsu\cite{Dasgupta2008HierarchicalSF} first proposed the idea of guided sampling by querying samples based on the results of hierarchical clustering. ALEC\cite{Wang2017ActiveLT} select representative samples drawn on the structure of the data.

Although a number of active methods query the label of a single sample, studies of active clustering prefer pairwise queries\cite{Chien2019HS2AL}.  Xiong et al.\cite{2013Spectral} proposed an active spectral clustering algorithm with k-nearest neighbor graphs, selecting pairwise constraints based on node uncertainty. Ashtiani et al. \cite{Ashtiani2016ClusteringWS} introduce a semi-supervised active clustering (SSAC) framework asking whether two given instances belong to the same cluster or not and demonstrate that access to simple query answers can turn an otherwise NP-hard clustering problem into a feasible one. Dasgupta and Ng\cite{Dasgupta2010WhichCD} poses the problem that clustering algorithms only group documents along the most prominent dimension without knowing the user’s intention, which is similar to our problem of diversity of image clustering criteria. It proposes an active spectral clustering algorithm, which makes it easy for a user to specify the dimension along which she wants to cluster the data points is sentiment.

\subsection{Deep Clustering}
Deep neural networks are explored to improve clustering performance due to their ability to learn representations on complex high-dimensional datasets\cite{Vincent2010StackedDA, Kingma2014AutoEncodingVB}. Recent works focus on end-to-end methods to transform the data into clustering-oriented representations. For example, DEC\cite{Xie2016UnsupervisedDE} optimizes the cluster centers and embedded features simultaneously by minimizing the KL-divergence for features in the latent subspace. DCN\cite{Yang2017TowardsKS} scatter samples in the low-dimensional space around their corresponding cluster centroids to learn a K-means friendly representation. IDFD\cite{TaoTN21} is a spectral clustering friendly representation learning by reducing correlations within features.

Self-augmentation based methods also achieve good performance. IIC\cite{Ji2019InvariantIC} maximizes the mutual information between positive instance pairs to discover clusters. PICA\cite{Huang2020DeepSC} clusters by minimizing the cosine similarity between the cluster-wise assignment vectors to learn the most semantically plausible clustering solution. DCCM\cite{Wu2019DeepCC} introduces the augmentation and utilizes the correlations among representations.
Inspired by the above ideas, CC\cite{Li2021ContrastiveC} proposes a dual contrastive learning framework. This method is based on the observation of ``label as representation",  conducting contrastive learning at not only the instance-level but also the cluster-level to learn clustering-favorite representations. GCC\cite{Zhong2021GraphCC} selects the positive pairs and negative pairs by the KNN graph constructed on the instance representation. %SCCL\cite{Zhang2021SupportingCW} is an extension of the contrastive clustering method on textual data, proving that this self-supervised framework is universally applicable.

These methods have achieved excellent results on large datasets in default orientation but cannot cluster personalized. Our approach introduces active learning into deep clustering, where oracles guide the network to learn cluster-oriented features.

\begin{figure*}[ht]
\centering
\includegraphics[width=0.95\textwidth]{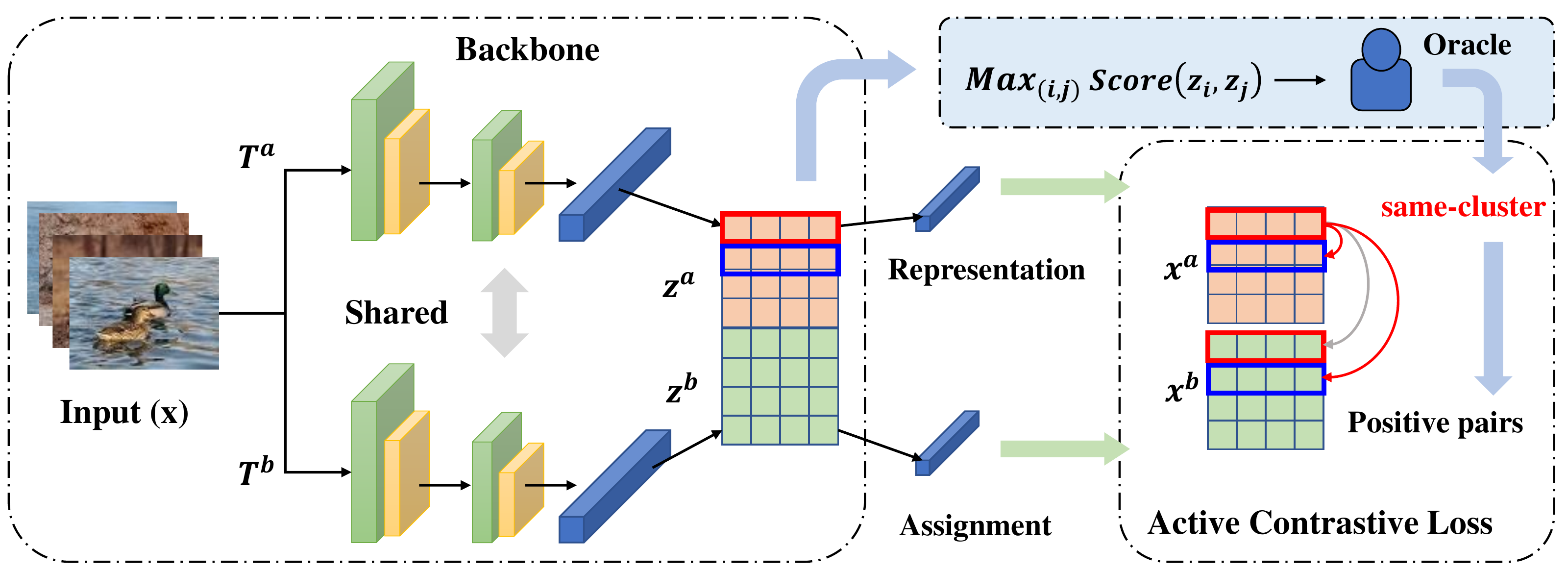} 
\caption{One active cycle of Oracle-guided Contrastive Clustering. A shared deep neural network generates representations from two random augmentations of the data, projected into representation space and assignment space. We explore a scoring function to measure the informativeness of each sample pair and select the largest one to submit a query to the oracle. The obtained same-cluster pairs are leveraged to extend the positive instance pairs in the contrastive loss.}
\label{fig2}
\end{figure*}

\section{Method}
\subsection{Proposed model: OCC}
We propose a novel model, named OCC, by exploiting active learning to guide the network to cluster in a given orientation and to improve clustering performance as well. One active cycle of OCC is illustrated in Fig. \ref{fig2}. Fig. \ref{fig2} depicts that active queries to oracles are embedded in the loop of OCC. In terms of the designed scoring function on the embedding features, some instance pairs are actively selected for annotation, and they are judged whether belonging to the same cluster in the forms of yes or no. Simultaneously, to learn cluster-friendly features and to compact cluster assignments, respectively, OCC projects the embedding features into representation and assignment spaces. Finally, these two spaces, augmented with annotated instance pairs, are optimized with bi-objectives: better clustering performance along a given orientation by leveraging active contrastive loss.

%in each active cycle of OCC, two random augmentations to each image is first constructed as an initial positive pair\cite{chen2020simple, Li2021ContrastiveC}. The features of the augmented samples are then extracted by leveraging a convolutional neural network. Third, some informative instance pairs are actively selected in terms of the designed scoring function for annotation, and they are judged whether belonging to the same cluster in forms of yes or no. The judgements are recorded in a matrix for clustering iterative updation. To learn cluster-friendly features and to compact cluster assignments, respectively, OCC projects the features into representation and assignment spaces. Finally, contrastive learning in both spaces, augmented with annotated instance pairs, is conducted for bi-objectives: better clustering performance along a given orientation. 

In OCC, the contrastive loss is recontructed by extending the positive instance pairs with pairwise annotations given by the oracle. For example, given a data $x_i$, we first select two stochastic data transformations $T^a$, $T^b$ from the same family of augmentations $\mathcal T$ and apply them to the data. Therefore, the augmented samples $x_i^a=T^a(x_i)$ and $x_i^b=T^b(x_i)$ constitute the initial positive sample pair $(x_i^a, x_i^b)$. At this time, the oracle informs that the samples of $x_i$ and $x_j$ belong to the same cluster, then the positive instances related to $x_i^a$ include $x_i^b$, $x_j^a$, and $x_j^b$.

In the followed two subsections, we describe the construction of our contrastive loss and query strategy in detail.

\subsection{Active Contrastive Loss}
Given a mini-batch size of $N$, let $X$ denotes the feature matrix of $2N$ augmented samples $\{x_1^a,...,x_N^a, x_1^b,...,x_N^b\}$ in the space. The same-cluster sample pairs given by the oracle is represented by a matrix $C_{N*N}$ and
\begin{equation}\label{eq:3}
C_{i,j} = \left\{
\begin{aligned}
\lambda,\quad &y_i = y_j \\
0,\quad &others
\end{aligned}
\right.
\end{equation}
where $\lambda$ is a variable weight parameter that controls the weight between the initial positive pair and the queried positive pair. $C_{i,j}$ indicates if the oracle annotations $x_i$ and $x_j$ belong to the same cluster. We leverage cosine similarity to measure the similarity between $x_i$ and $x_j$, i.e.
\begin{equation}\label{eq:4}
sim(x_i,x_j)= \frac{(x_i)(x_j)^\top}{\lVert x_i\rVert \lVert x_j\rVert}
\end{equation}
Let $s(x_i,x_j)=exp(sim(x_i,x_j)/\tau)$, where $\tau$ denotes a temperature parameter. For an augmented sample $x_i^a$, the relevant positive instances are $x_i^b$ and $x_j^a$, $x_j^b$ if $C_{i,j}>0$. To narrow the distance between positive instances pairs and learn their similar features, we set the objective function to be the ratio of the similarity between positive pairs to the similarity between negative pairs. The active contrastive loss of $x_i^a$ is defined as
\begin{equation}\label{eq:5}
l_i^a=\frac{s(x_i^a, x_i^b)+\sum_{j=1}^{N}C_{i,j}[s(x_i^a, x_j^a)+s(x_i^a, x_j^b)]}{\sum_{j=1}^{N}(C_{i,j}+1)\sum_{j=1}^{N}[s(x_i^a, x_j^a)+s(x_i^a, x_j^b)]}
\end{equation}
where $\sum_{j=0}^{N}(C_{i,j}+1)$ is a method of data normalization to control the range of the loss. In the denominator, we leverage all of the samples instead of the negative instances for the same reason. Although it seems to affect the similarity to all instances, including itself, it only reduces the distance between negative instances due to the increased distance between positive instances in the molecule.

The loss in the sample space $X$ is the sum of all sample losses, i.e.
\begin{equation}\label{eq:6}
\mathcal L(X,C)=\frac{1}{2N}\sum_{i=1}^{N}(-log(l_i^a)-log(l_i^b))
\end{equation}

The network extracts features $Z$ from the augmented samples and projects the features into the representation space $\hat{Z}\in\mathcal{R}^{2N*M}$ and assignment space $\hat{Y}\in\mathcal{R}^{2N*K}$, where $M$ is a preset feature dimension and $K$ is the number of clusters. $\hat{Z}_i$ is the $M$-dimensional feature of the augmented sample $x_i$ and $\hat{Y}_i$ is the assignment probability vector of the clusters of $x_i$. In general, contrastive methods learn the similar features of positive instance pairs in the representation space and narrow the intra-cluster distance while expanding the inter-cluster distance in the assignment space. Here we also implement contrastive learning of instances in the assignment space though these two spaces are related to a certain extent. It will encourage the positive pairs to be allocated in the same cluster and achieve better performance in practice.
The overall objective function is defined as
\begin{equation}\label{eq:7}
\mathcal L=\mathcal L(\hat{Z},C)+\mathcal L(\hat{Y},C)+\mathcal L(\hat{Y}^\top,O)-H(\hat{Y})
\end{equation}
where $C$ is the query matrix defined in Eq. \ref{eq:3} so that $\mathcal L(\hat{Z},C)+\mathcal L(\hat{Y},C)$ denotes the contrastive loss for instances on representation space and assignment space. Matrix $O$ is a zero matrix and $\mathcal L(\hat{Y}^\top,O)$ denotes the contrastive loss for clusters on assignment space. According to Eq. \ref{eq:5}, the active loss function degenerates into the initial contrastive loss if $C=O$. Positive pairs at the clustering level consist only of the same column of the cluster assignment matrix of the augmented samples, so the contrastive loss is not expanded. $H(\hat{Y})=\sum_{i=1}^KP(\hat{y}_i)logP(\hat{y}_i)$ where $P(\hat{y}_i)=\sum_{j=1}^{2N}\hat{Y}_{j,i}/||\hat{Y}||_1$, which is leveraged to balance the number of samples in each cluster.

\begin{algorithm}[tb]
\caption{Oracle-guided Contrastive Clustering}
\label{alg:algorithm}
\textbf{Input}: Training dataset $\mathcal X$; Cluster number $K$; Training epochs $E$; Batch size $N$.\\
\textbf{Parameter}: Query times $Q$; Temperature parameter $\tau$; Hyper-parameters $\lambda$.\\
\textbf{Output}: The clustering result $\mathcal C$.
\begin{algorithmic}[1] %[1] enables line numbers
\FOR{$epoch=1$ to $E$}
\FOR{a sampled mini-batch $\{x_i\}_{i=1}^N$}
\STATE Generating augmentations for the sampled images;
\STATE Utilize the network to extract feature matrix $Z$;
\STATE Select the sample pair with the highest score according to Eq. \ref{eq:8};
\STATE The oracle indicates the sample pair and records it with Eq. \ref{eq:3};
\STATE Project $Z$ onto the representation space to get $\hat{Z}$;
\STATE Project $Z$ onto the assignment space to get $\hat{Y}$;
\STATE Calculate the active contrastive loss $\mathcal L$ through Eq. \ref{eq:4}--\ref{eq:7};
\STATE Update the network by minimizing $\mathcal L$;
\ENDFOR
\ENDFOR
\STATE Calculate $\mathcal C=argmax(\hat{Y})$
\STATE \textbf{return} $\mathcal C$
\end{algorithmic}
\end{algorithm}

\subsection{Pairwise Query Strategy}
In the pairwise query strategy, oracles judge whether the sample pairs belong to the same cluster to get pairwise constraints. Thus the network clusters along to the desired orientation and achieves better clustering performance. The scoring function Cyclic Similarity Discrepancy(CSD) of a sample pair $(x_i,x_j)$ is defined as
\begin{equation}\label{eq:8}
score(x_i,x_j)=s_c(x_i,x_j)|s_c(x_i,x_j)-s_{c-1}(x_i,x_j)|
\end{equation}
where $s_c(x_i,x_j)=sim(x_i,x_j|c)$ denotes the similarity of sample pair $(x_i,x_j)$ in the $c-th$ iteration. The method is inspired by TOD\cite{Huang2021SemiSupervisedAL}, which argues that samples with the most significant change in features also have the highest true loss. We tend to select sample pairs that are likely to be similar and have a significant discrepancy in the similarity during the iterative process. Such samples have been shown to have large losses in TOD. In the following section, we will prove that selecting these samples helps to receive a tighter bound of clustering risk.

\subsection{Generalization bound}
Clustering aims to divide the samples into several clusters
such that samples lying in the same cluster have more similarities than those in others. We formally define the problem of clustering as minimizing the following criterion:
\begin{equation}
\begin{aligned}
\mathbb{E}_{z\sim \mathcal{Z}}\left[l\left(z ; h\right)\right] = \mathbb{E}_{z\sim \mathcal{Z}}[1-s(z_{i},z_{j}; h)]
\end{aligned}
\label{eq:1}
\end{equation}
where $Z$ is the population of same-cluster pairs, $s$ is a function measures the similarity of two samples in sample pair $z$. This criterion is expected clustering risk proposed by Liu et al.\cite{li2021sharper}. %Clustering framework of minimizing expected clustering risk is generalized well and suitable for a lot of clustering algorithms. 

The optimization objective we define in Eq. \ref{eq:1} is not directly computable since we do not have access to all the information of sample pairs. In order to design an active learning strategy which is effective in pairwise query setting, we consider the following decomposition of expected clustering risk. This is a probabilistic sampling procedure inspired from Pydi et al.\cite{pydi2019active}. 
\begin{equation}
\begin{aligned}
\mathbb{E}_{z\sim \mathcal{Z}}\left[l\left(z ; h_{S}\right)\right] 
&\le \underbrace{\left.\mid \mathbb{E}_{z\sim \mathcal{Z}}\left[l\left(z, h_{S}\right)\right)\right]-\frac{1}{|T|} \sum_{z\in T} l\left(z, h_{S}\right) \mid}_{(A)\quad excess\quad clustering \quad risk}  \\
&+\underbrace{\frac{1}{|T|} \sum_{z\in T} \frac{Q_{z}}{p_{z}} l\left(z, h_{S}\right)}_{\quad\quad(B) \quad extended \quad clustering \quad risk} \\ &+\underbrace{|\frac{1}{|T|} \sum_{z\in T} l\left(z, h_{S}\right)-\frac{1}{|T|} \sum_{z\in T}
\frac{Q_{z}}{p_{z}} l\left(z, h_{S}\right)|}_{(C)\quad active \quad clustering \quad risk}
%(B) \quad weighted \quad training \quad error
%(C)\quad active \quad learning \quad error
\end{aligned}
\label{eq:2}
\end{equation}
where $T$ denotes the same-cluster pairs in our samples defined as target sample pairs, $p_{z}$ denotes the probabilities of sample pairs $z$ being queried, $Q_{z}$ denotes a set of Bernoulli random variables such that $P(Q_{z} = 1) = p_{z}$. Sample pairs $z$ will be queried when $Q_{z} = 1$.

Term (A) corresponds to the excess clustering risk of the algorithm $h_{S}$. It denotes the difference between expected clustering risk and empirical clustering risk. (B) corresponds to active clustering risk for $h_{S}$ over the sample pairs queried. This term could be minimized directly during the training process. (C) corresponds to active clustering risk. It is the absolute difference
between the average clustering risk over all target sample pairs and the active clustering risk of
the queried sample pairs.

According to Liu et al.\cite{li2021sharper}, the excess clustering risk bounds in Eq. \ref{eq:2} are mostly of order $\mathcal{O}(\sqrt{K} / \sqrt{n})$ provided
that the underlying distribution has bounded support. In a large data set, $n$ is much more large than $K$, term (A) could be small. Moreover, it is widely observed that the deep neural networks are highly expressive leading to very low training risk on selected sample pairs. Empirically, the active clustering risk is small. Hence, the critical part for active clustering are active clustering risk. The following theorem is presented to analyze its upper bound.

\subsubsection{Theorem 1.} \textit{Define $D_{p}:=\sum_{z \in T} \frac{l(z, h_{S})}{p_{z}}$. Let $c_{\delta}>0$ be a constant that depends on $\delta$. Active clustering risk can be bounded as follows, with probability at least $1-\delta$}
\begin{equation}\label{eq:9}
\begin{aligned}
&|\frac{1}{|T|} \sum_{z\in T} l\left(z, h_{S}\right) -\frac{1}{|T|} \sum_{z\in T}
\frac{Q_{z}}{p_{z}} l\left(z, h_{S}\right)| \le c_{\delta} \frac{D_{p}}{|T|}
\end{aligned}
\end{equation}

\noindent Proof of Theorem 1. Define $H_{z}=l(z,h_{S})-\frac{Q_{z}}{p_{z}} l\left(z, h_{S}\right)$, $M=\sum_{z\in T} E[H_{z}^{2}]$, we get the following bound on $M$
\begin{equation}\label{eq:10}
\begin{aligned}
M
&=\sum_{z\in T} Var[H_{z}^{2}] \\
&=\sum_{z\in T} l(z,h_{S})^2(\frac{1}{p_{z}}-1) \\
&\le \sum_{z\in T} \frac{l(z,h_{S})^2}{p_{z}^2}=D_{p} \\
\end{aligned}
\end{equation}

We further use the Bernstein’s inequality and conclude that with probability at least $1-\delta$
\begin{equation}\label{eq:11}
\begin{aligned}
&|\frac{1}{|T|} \sum_{z\in T} l\left(z, h_{S}\right)-\frac{1}{|T|} \sum_{z\in T}\frac{Q_{z}}{p_{z}} l\left(z, h_{S}\right) | \\
&=\frac{1}{|T|}\sum_{z\in T}H_{z} \\
&\leq \frac{D_{p}}{3 |T|} \log \frac{1}{\delta}\left(1+\sqrt{1+\frac{18}{\log \frac{1}{\delta}}}\right)
\end{aligned}
\end{equation}
\qed

Unlike the one proposed by Pydi et al. \cite{pydi2019active}, our bound is tighter because we choose true loss instead of pseudo-loss. It is clear from Eq. \ref{eq:9} that choosing $p$ so as to minimize $D_{p}$ will result in the tightest bound for the expected clustering loss. In the next theorem, we present the optimal sampling probability distribution $p^{*}$ that minimizes $D_{p}$.

\subsubsection{Theorem 2.} \textit{The optimal distribution $p^{*}$ for minimizing $D_{p}:=\sum_{z \in T} \frac{l(z, h_{S})}{p_{z}}$ is given by}
$$
p_{z}^{*}=\frac{l(z, h_{S})^{1 / 2}}{\sum_{z \in T} l(z, h_{S})^{1 / 2}}
$$
Proof of Theorem 2 is the same with proof of Theorem 6.3 in  Pydi et al.\cite{pydi2019active}. 

This theorem suggests that to minimize the clustering risk, it is workable to design a query strategy by selecting instance pairs with higher true loss. As the true losses of the instance pairs are distinct under various clustering criteria, the instance pairs with high cluster loss indicate that they are key instances that signify clustering tasks. Our pairwise query strategy is a CSD-based data sampling strategy, theoretically, instance pairs with large clustering risk can be obtained in the unlabeled pool, so as to minimize the expected clustering risk by active learning. 

\section{Experiments}
\subsection{Experimental Settings}
\subsubsection{Datasets}
To evaluate the effectiveness of the proposed method, we
conduct experiments on four widely-used image datasets,
including CIFAR-10, CIFAR-100\cite{Krizhevsky2009LearningML}, ImageNet-10 and ImageNet-Dogs\cite{Chang2017DeepAI}. We use the training and test sets of CIFAR-10 and CIFAR-100, and only make use of the training set of ImageNet-10 and ImageNet-Dogs.

Artificially, we divide the initial classes of the dataset into several target-clusters according to two distinct demands to simulate the diversity of clustering orientation. As depicted in Fig. \ref{fig1}, two distinct oracle demands cluster the images along distinct orientations. It is worth noting that not every demand matches the semantics of reality. Tab. \ref{table:1} illustrates the details of the adopted datasets.

\begin{table}[tb]   
\begin{center}
\begin{tabular}{cccc}   
\toprule  
\textbf{Dataset} & \textbf{Samples} & \textbf{Classes} & \textbf{Target Clusters} \\   
\midrule   
CIFAR-10 & 60,000 & 10 & 2  \\   
CIFAR-100 & 60,000 & 100 & 4  \\   
ImageNet-10  & 13,000 & 10 & 2  \\   
ImageNet-Dogs  & 19,500 & 15 & 2  \\   
\bottomrule   
\end{tabular}   
\caption{A summary of the datasets.}  
\label{table:1} 
\end{center}   
\end{table}

\begin{table*}[ht]
  \centering
    \begin{tabular}{ccccccccccccc}
    \toprule 
          & \multicolumn{3}{c}{CIFAR-10} & \multicolumn{3}{c}{CIFAR-100} & \multicolumn{3}{c}{ImageNet-10} & \multicolumn{3}{c}{ImageNet-dogs} \\
    \midrule
          \textbf{Default} & NMI   & ARI   & ACC   & NMI   & ARI   & ACC   & NMI   & ARI   & ACC   & NMI   & ARI   & ACC \\
    \midrule
    K-means & 0.034 & 0.063 & 0.629 & 0.043 & 0.038 & 0.356 & 0.007 & 0.008 & 0.544 & 0.000 & -0.002 & 0.518 \\
    PICA  & 0.550  & 0.629 & 0.897 & 0.15  & 0.157 & 0.439 & 0.887 & 0.939 & 0.985 & 0.207 & 0.234 & 0.742 \\
    GCC   & 0.675 & 0.763 & 0.937 & 0.269 & 0.259 & 0.544 & 0.897 & 0.947 & 0.986 & 0.233 & 0.313 & 0.78 \\
    CC    & 0.602 & 0.646 & 0.902 & 0.369 & 0.385 & 0.701 & \textbf{0.924} & \textbf{0.963} & \textbf{0.991} & 0.179 & 0.221 & 0.735 \\
    IDFD  & 0.879 & 0.937 & 0.984 & 0.329 & 0.221 & 0.511 & 0.923 & 0.962 & 0.99  & 0.338 & 0.391 & 0.815 \\
    OCC   & \textbf{0.884} & \textbf{0.939} & \textbf{0.985} & \textbf{0.479} & \textbf{0.512} & \textbf{0.782} & 0.918 & 0.959 & 0.990  & \textbf{0.512} & \textbf{0.602} & \textbf{0.888} \\
    \bottomrule
          &       &       &       &       &       &       &       &       &       &       &       &  \\
    \toprule
          \textbf{Personalized} & NMI   & ARI   & ACC   & NMI   & ARI   & ACC   & NMI   & ARI   & ACC   & NMI   & ARI   & ACC \\
    \midrule
    K-means & 0.002 & 0.003 & 0.503 & 0.046 & 0.049 & 0.362 & 0.003 & 0.004 & 0.533 & 0.000     & -0.000 & 0.506 \\
    PICA  & 0.000     & 0.001 & 0.517 & 0.250  & 0.253 & 0.587 & 0.025 & 0.034 & 0.593 & 0.032 & 0.041 & 0.601 \\
    GCC   & 0.037 & 0.036 & 0.616 & 0.283 & 0.267 & 0.582 & 0.023 & 0.031 & 0.588 & 0.063 & 0.089 & 0.649 \\
    CC    & 0.006 & 0.009 & 0.546 & 0.214 & 0.202 & 0.485 & 0.025 & 0.035 & 0.594 & 0.035 & 0.047 & 0.609 \\
    IDFD  & 0.021 & 0.038 & 0.599 & 0.175 & 0.095 & 0.469 & 0.025 & 0.035 & 0.593 & 0.280  & 0.266 & 0.758 \\
    OCC   & \textbf{0.690} & \textbf{0.791} & \textbf{0.945} & \textbf{0.538} & \textbf{0.569} & \textbf{0.801} & \textbf{0.732} & \textbf{0.810} & \textbf{0.950} & \textbf{0.702} & \textbf{0.801} & \textbf{0.948} \\
    \bottomrule
    \end{tabular}%
  \caption{The clustering performance under two clustering orientations, default and personalized, on four object image benchmarks. ‘default’ perform clustering along the default orientation, while ‘personalized’ along a given and personalized orientation. The best results are shown in boldface.}
  \label{table:2}%
\end{table*}%

\subsubsection{Implementation Details}
For fairness, ResNet34 is adopted as the backbone network  without any modification. The parameters related to deep contrastive clustering are set following previous methods\cite{Li2021ContrastiveC, Zhong2021GraphCC}. Adam with an initial learning rate of 0.0003 is adopted to optimize the network and  the batch size is set as 256. In addition, We resize all images uniformly to the size of $224\times224$. The feature dimensionality $M$ of the instances in representation space is set to 128. 

 Similar to the decay mechanism, the parameter $\lambda$ in Eq. \ref{eq:3} is set to $(1000-e)*0.05$ where $e$ is the current epoch num. Thus, the proportion of each positive pair gradually decreases as the number of queries increases. We query about 25\% of the instance pairs for each dataset and twice per batch. Moreover, after a number of iterations, we extend annotation by pseudo labeling the instances similar to annotated ones with a high confidence.

%To ensure fairness in comparison, we adopt ResNet34 as the backbone network without any modification. The settings of parameters related to deep contrastive clustering is consistent with previous methods\cite{Li2021ContrastiveC, Zhong2021GraphCC}. Furthormore, the images are resized uniformly to the size of $224\times224$, and the feature dimensionality $M$ of the samples in representation space is set to 128. We adopt the Adam optimizer with an initial learning rate of 0.0003 to optimize the network and set the batch size to 256 due to the memory limitation.

%The parameter $\lambda$ in Eq. \ref{eq:3} is constructed to $(1000-e)*0.05$ where $e$ is the current epoch num. This setting is similar to the decay mechanism. When we only have a small number of labeled instance pairs, we let active learning occupy a large ratio, and the proportion of each positive pair gradually decreases as the number of queries increases. We query about 25\% of the sample pairs for each dataset and twice per batch. In addition, we also perform simple extension labeling and select sample pairs with high similarity and confidence to label pseudo-labels(same-cluster) after training for a certain period of time.

All comparative experiments are implemented with three NVIDIA TU102 RTX 2080  Ti GPUs on PyTorch platform.

\subsubsection{Compared Methods}
We compared the proposed method with both traditional and deep learning based methods, including K-means\cite{MacQueen1967SomeMF}, PICA\cite{Huang2020DeepSC}, GCC\cite{Zhong2021GraphCC}, CC\cite{Li2021ContrastiveC}, IDFD\cite{TaoTN21}. We set the target number of clusters to the number of target clusters for all methods and run them in a unified environment.
Moreover, two known active query strategies are compared with our adopted strategy in our designed deep framework.

\subsubsection{Evaluation Metrics}
We adopted three standard clustering metrics to evaluate our method including Normalized Mutual Information (NMI), Accuracy (ACC), and Adjusted Rand Index (ARI). These metrics reflect the performance of clustering from different aspects, and higher values indicate better performance. 

\subsection{Experimental Results}
\subsubsection{Clustering Performance}
We presented the clustering performance of OCC and the compared methods on four datasets with two distinct cluster orientations in Tab. \ref{table:2}. From Table 2, we observe the following facts. i) In the default clustering orientation,, the clustering performance of our OCC is obviously better than the comparison deep clustering algorithms in most.  It is observed that on ImageNet-10, only IDFD slightly performs better than OCC.  than . These shows that, without personalized clustering demands, our OCC can outperform the compared SOTA clustering methods. ii) In the personalized clustering orientation, the performance of the SOTA clustering methods is relatively low. This shows that the existing SOTA clustering methods are unworkable and not applicable for the clustering tasks with personalized demands. iii) In the personalized clustering orientation, the performance of OCC is high and extremely better than the compared SOTA clustering methods. This shows that unlike the existing methods, OCC is apt to clustering with personalized demands. it also shows that the active query is workable to guide clustering along a given orientation. 

\begin{figure}[t]
\centering
\includegraphics[width=0.9\columnwidth]{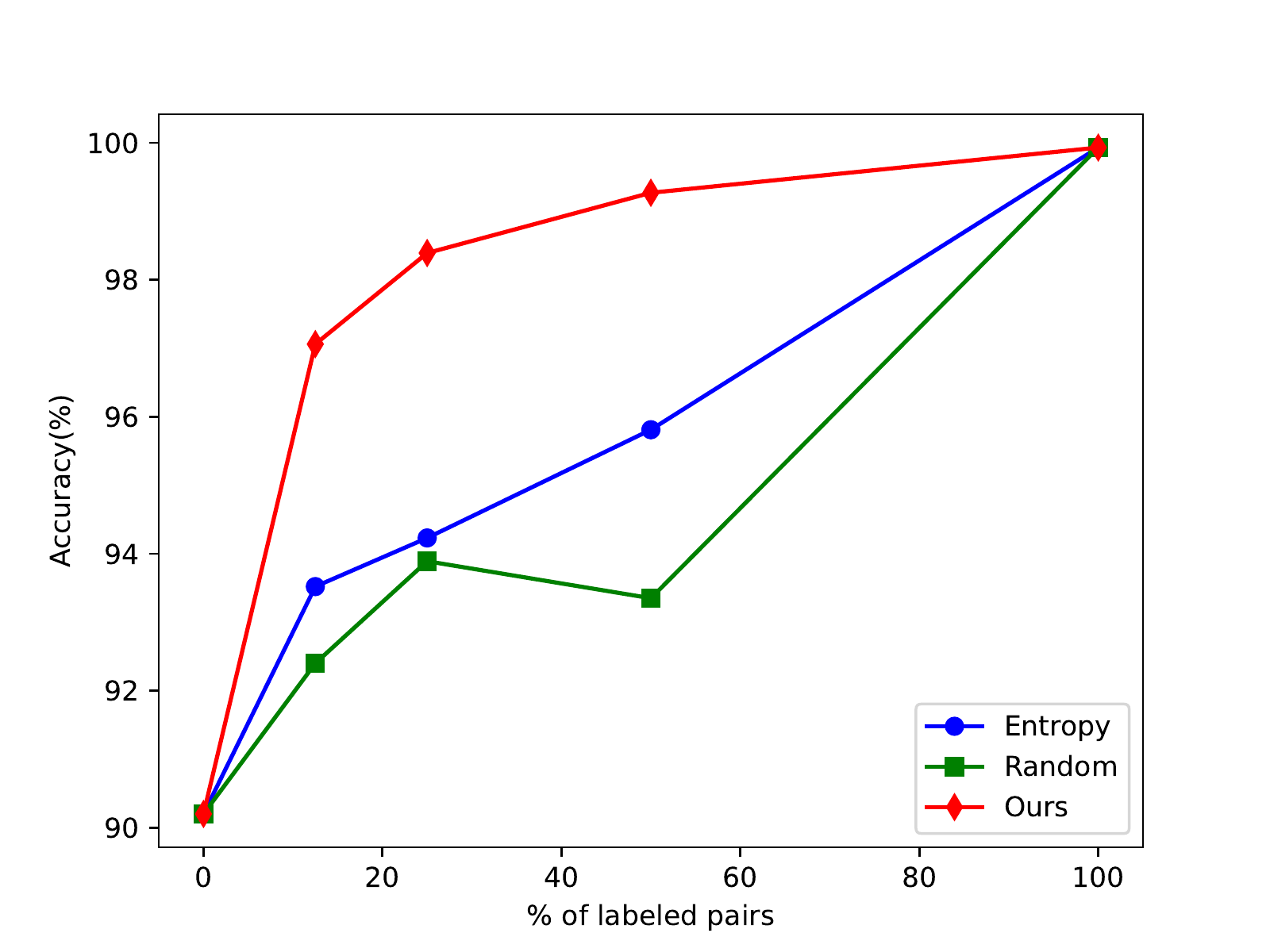}
\caption{The accuracy of three query strategies available on CIFAR-10.}.
\label{fig3}
\end{figure}

\subsubsection{Query efficiency}
To verify the effectiveness of the query strategy, we compare our adopted query strategy with two known ones, including entropy-based query and random query, in our designed deep clustering framework. The random query strategy is to select instance pairs randomly. The entropy-based query strategy selects the instance with the maximum entropy and paired it with another medium similar one. The results of the comparison on CIFAR-10 are shown in Fig. \ref{fig3}. The percentage of queried ones to the total instance pairs (about 200,000 for CIFAR-10) is seen as the query cost. And it is headed as ‘\% of labeled pairs’ as the title of the vertical axis in Fig. \ref{fig3} . Note that the results of unsupervised(0\%) and full-supervised(100\%) learning are independent from query strategy.

We have the following observations from Fig. \ref{fig3}. i) it is observed that the red trendline is always higher than the blue and green ones. This indicates that our adopted query strategy outperforms these two known active query ones. ii) At 25\% of the query cost, our query strategy almost reaches the highest accuracy. This shows that OCC saves a lot of annotation cost. 

\begin{figure}[t]
\centering
\includegraphics[width=0.95\columnwidth]{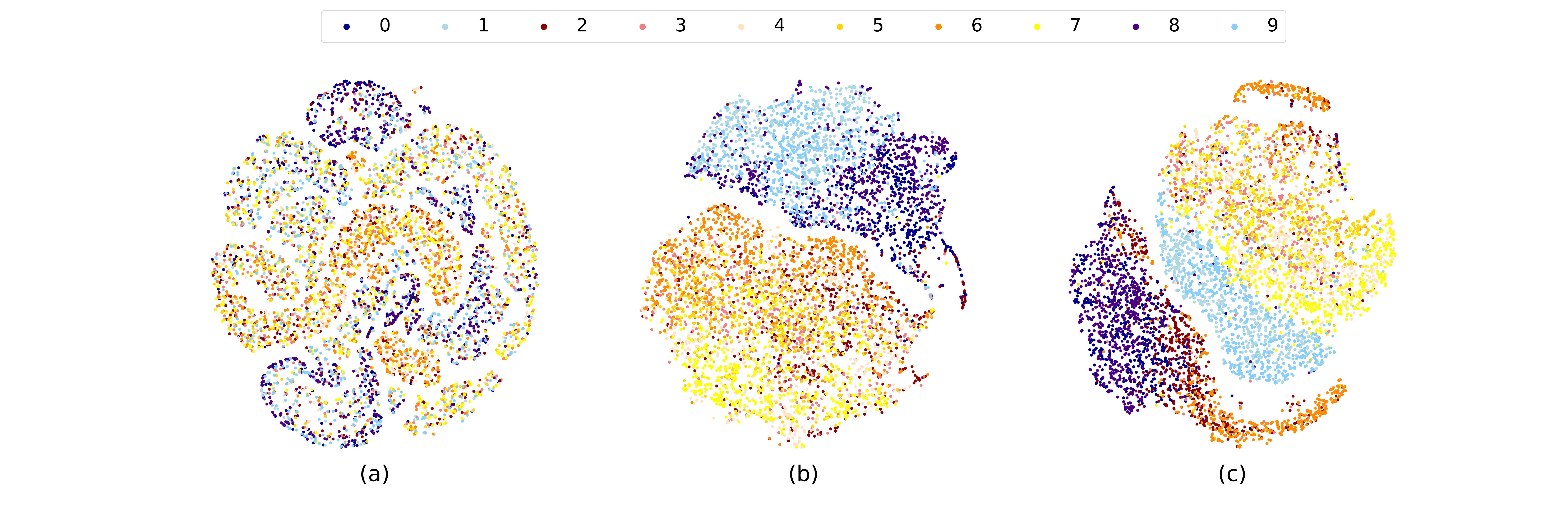} 
\caption{The evolution of features across the training process under two clustering orientations: (a) Initial distribution of instances, (b) Instance distribution after clustering along the default orientation and (c) Instance distribution after clustering along a given personalized orientation. The colors of the dots denote the 10 class labels of CIFAR-10.}
\label{fig4}
\end{figure}

\subsection{Visualization}
To vividly display the personalized clustering results, we visualize the distribution of instances in CIFAR-10 after clustering. In Fig. \ref{fig4}, we have the following observations. i) Subfigure (a) shows that the instance points before clustering is chaos. ii) Subfigure (b) shows that the yellow and orange instance points cluster together, while in subfigure (c) the dark blue and dark orange dots are clustered together. This shows that OCC realizes personalized clusters. 

\subsection{Ablation Studies}
\subsubsection{Effect of contrastive learning}
We perform ablation analysis by removing the contrastive part of each one of the representation and assignment spaces. Along both default and personalized clustering orientation of CIFAR-10, the results are shown in Tab. \ref{table:3}. We observe that the clustering performance with contrastive learning in both spaces always obtains the highest values. This shows it is necessary to conduct contrastive learning in both spaces and it is effective in accurately catching the desired features by contrastive learning. 

\begin{table}[tb]   
\begin{center}
\begin{tabular}{ccccc}   
\toprule  
\textbf{Orientation} & \textbf{Contrastive space} & \textbf{NMI} & \textbf{ARI} & \textbf{ACC} \\   
\midrule   
\multirow{3}{*}{Default} 
& R+A & 0.884 & 0.939 & 0.985  \\
& R only & 0.631 & 0.683 & 0.913  \\
& A only & 0.589 & 0.628 & 0.896  \\ 
\midrule   
\multirow{3}{*}{Personalized} 
& R+A & 0.690 & 0.791 & 0.945  \\
& R only & 0.005 & 0.006 & 0.539  \\
& A only & 0.492 & 0.499 & 0.853  \\ 
\bottomrule   
\end{tabular}   
\caption{Effect of contrastive learning in representation(R) and assignment(A) space on CIFAR-10.} 
\label{table:3}  
\end{center}   
\end{table}

\subsubsection{Effect of label extension}
We perform ablation analysis by removing the operation of label extension. Along both default and personalized clustering orientation of CIFAR-10, the results are shown in Tab. \ref{table:4}. It is observed that it can obtain a substantially higher performance with label extension. This shows that label extension benefits clustering performance improvement by increasing the annotated instance pairs.

\begin{table}[tb]   
\begin{center}
\begin{tabular}{ccccc}   
\toprule  
\textbf{Orientation} & \textbf{Label Extension} & \textbf{NMI} & \textbf{ARI} & \textbf{ACC} \\   
\midrule   
\multirow{2}{*}{Default} 
& YES & 0.884 & 0.939 & 0.985  \\
& NO & 0.631 & 0.683 & 0.913  \\
\midrule   
\multirow{2}{*}{Personalized} 
& YES & 0.690 & 0.791 & 0.945  \\
& NO & 0.524 & 0.542 & 0.868  \\
\bottomrule   
\end{tabular}   
\caption{Effect of label extension on CIFAR-10.} 
\label{table:4}  
\end{center}   
\end{table}

\section{Conclusion}
We have presented a model, named Oracle-guided Contrastive Clustering(OCC), for task-aware clustering that incorporate active query to oracles with personalized demand. Moreover, contrastive learning is joint with active learning to catch orientation-aware features and achieve desired clustering solutions. This is first deep framework for personalized image clustering. In the near future, we would like to extend this framework to any clustering domain in general.
%\nobibliography{aaai23}

%\bibliography{main.bbl}

% Generated by IEEEtran.bst, version: 1.14 (2015/08/26)
\begin{thebibliography}{10}
\providecommand{\url}[1]{#1}
\csname url@samestyle\endcsname
\providecommand{\newblock}{\relax}
\providecommand{\bibinfo}[2]{#2}
\providecommand{\BIBentrySTDinterwordspacing}{\spaceskip=0pt\relax}
\providecommand{\BIBentryALTinterwordstretchfactor}{4}
\providecommand{\BIBentryALTinterwordspacing}{\spaceskip=\fontdimen2\font plus
\BIBentryALTinterwordstretchfactor\fontdimen3\font minus
  \fontdimen4\font\relax}
\providecommand{\BIBforeignlanguage}[2]{{%
\expandafter\ifx\csname l@#1\endcsname\relax
\typeout{** WARNING: IEEEtran.bst: No hyphenation pattern has been}%
\typeout{** loaded for the language `#1'. Using the pattern for}%
\typeout{** the default language instead.}%
\else
\language=\csname l@#1\endcsname
\fi
#2}}
\providecommand{\BIBdecl}{\relax}
\BIBdecl

\bibitem{Xu2015ACS}
D.~Xu and Y.~Tian, ``A comprehensive survey of clustering algorithms,''
  \emph{Annals of Data Science}, vol.~2, pp. 165--193, 2015.

\bibitem{MacQueen1967SomeMF}
J.~MacQueen, ``Some methods for classification and analysis of multivariate
  observations,'' in \emph{Fifth Berkeley Symposium on Mathematical Statistics
  and Probability}, 1967, p. 281–297.

\bibitem{Zhang1996BIRCHAE}
T.~Zhang, R.~Ramakrishnan, and M.~Livny, ``Birch: an efficient data clustering
  method for very large databases,'' in \emph{SIGMOD '96}, 1996.

\bibitem{Bezdek1984FCMTF}
J.~C. Bezdek, R.~Ehrlich, and W.~E. Full, ``Fcm: The fuzzy c-means clustering
  algorithm,'' \emph{Computers \& Geosciences}, vol.~10, pp. 191--203, 1984.

\bibitem{Ester1996ADA}
M.~Ester, H.-P. Kriegel, J.~Sander, and X.~Xu, ``A density-based algorithm for
  discovering clusters in large spatial databases with noise,'' in \emph{KDD},
  1996.

\bibitem{Rasmussen1999TheIG}
C.~E. Rasmussen, ``The infinite gaussian mixture model,'' in \emph{NIPS}, 1999.

\bibitem{Min2018ASO}
E.~Min, X.~Guo, Q.~Liu, G.~Zhang, J.~Cui, and J.~Long, ``A survey of clustering
  with deep learning: From the perspective of network architecture,''
  \emph{IEEE Access}, vol.~6, pp. 39\,501--39\,514, 2018.

\bibitem{Caron2018DeepCF}
M.~Caron, P.~Bojanowski, A.~Joulin, and M.~Douze, ``Deep clustering for
  unsupervised learning of visual features,'' in \emph{ECCV}, 2018.

\bibitem{Yang2017TowardsKS}
B.~Yang, X.~Fu, N.~Sidiropoulos, and M.~Hong, ``Towards k-means-friendly
  spaces: Simultaneous deep learning and clustering,'' in \emph{ICML}, 2017.

\bibitem{Li2021ContrastiveC}
\BIBentryALTinterwordspacing
Y.~Li, P.~Hu, Z.~Liu, D.~Peng, J.~T. Zhou, and X.~Peng, ``Contrastive
  clustering,'' \emph{Proceedings of the AAAI Conference on Artificial
  Intelligence}, vol.~35, no.~10, pp. 8547--8555, May 2021. [Online].
  Available: \url{https://ojs.aaai.org/index.php/AAAI/article/view/17037}
\BIBentrySTDinterwordspacing

\bibitem{Liu2022DeepGC}
Y.~Liu, W.~Tu, S.~Zhou, X.~Liu, L.~Song, X.~Yang, and E.~Zhu, ``Deep graph
  clustering via dual correlation reduction,'' in \emph{AAAI}, 2022.

\bibitem{Ren2022ASO}
P.~Ren, Y.~Xiao, X.~Chang, P.~Huang, Z.~Li, X.~Chen, and X.~Wang, ``A survey of
  deep active learning,'' \emph{ACM Computing Surveys (CSUR)}, vol.~54, pp.
  1--40, 2022.

\bibitem{Beluch2018ThePO}
W.~H. Beluch, T.~Genewein, A.~N{\"u}rnberger, and J.~M. K{\"o}hler, ``The power
  of ensembles for active learning in image classification,'' \emph{2018
  IEEE/CVF Conference on Computer Vision and Pattern Recognition}, pp.
  9368--9377, 2018.

\bibitem{Ranganathan2017DeepAL}
H.~Ranganathan, H.~Venkateswara, S.~Chakraborty, and S.~Panchanathan, ``Deep
  active learning for image classification,'' \emph{2017 IEEE International
  Conference on Image Processing (ICIP)}, pp. 3934--3938, 2017.

\bibitem{Gal2017DeepBA}
\BIBentryALTinterwordspacing
Y.~Gal, R.~Islam, and Z.~Ghahramani, ``Deep bayesian active learning with image
  data,'' in \emph{Proceedings of the 34th International Conference on Machine
  Learning, {ICML} 2017, Sydney, NSW, Australia, 6-11 August 2017}, ser.
  Proceedings of Machine Learning Research, vol.~70.\hskip 1em plus 0.5em minus
  0.4em\relax {PMLR}, 2017, pp. 1183--1192. [Online]. Available:
  \url{http://proceedings.mlr.press/v70/gal17a.html}
\BIBentrySTDinterwordspacing

\bibitem{Ash2020DeepBA}
J.~T. Ash, C.~Zhang, A.~Krishnamurthy, J.~Langford, and A.~Agarwal, ``Deep
  batch active learning by diverse, uncertain gradient lower bounds,''
  \emph{ArXiv}, vol. abs/1906.03671, 2020.

\bibitem{sener2018active}
\BIBentryALTinterwordspacing
O.~Sener and S.~Savarese, ``Active learning for convolutional neural networks:
  A core-set approach,'' in \emph{International Conference on Learning
  Representations}, 2018. [Online]. Available:
  \url{https://openreview.net/forum?id=H1aIuk-RW}
\BIBentrySTDinterwordspacing

\bibitem{Huang2021SemiSupervisedAL}
S.~Huang, T.~Wang, H.~Xiong, J.~Huan, and D.~Dou, ``Semi-supervised active
  learning with temporal output discrepancy,'' \emph{2021 IEEE/CVF
  International Conference on Computer Vision (ICCV)}, pp. 3427--3436, 2021.

\bibitem{Ashtiani2016ClusteringWS}
\BIBentryALTinterwordspacing
H.~Ashtiani, S.~Kushagra, and S.~Ben-David, ``Clustering with same-cluster
  queries,'' in \emph{Advances in Neural Information Processing Systems},
  vol.~29.\hskip 1em plus 0.5em minus 0.4em\relax Curran Associates, Inc.,
  2016. [Online]. Available:
  \url{https://proceedings.neurips.cc/paper/2016/file/9597353e41e6957b5e7aa79214fcb256-Paper.pdf}
\BIBentrySTDinterwordspacing

\bibitem{Dasgupta2010WhichCD}
S.~Dasgupta and V.~Ng, ``Which clustering do you want? inducing your ideal
  clustering with minimal feedback,'' \emph{J. Artif. Intell. Res.}, vol.~39,
  pp. 581--632, 2010.

\bibitem{Zhong2021GraphCC}
H.~Zhong, J.~Wu, C.~Chen, J.~Huang, M.~Deng, L.~Nie, Z.~Lin, and X.~Hua,
  ``Graph contrastive clustering,'' \emph{2021 IEEE/CVF International
  Conference on Computer Vision (ICCV)}, pp. 9204--9213, 2021.

\bibitem{Vincent2010StackedDA}
P.~Vincent, H.~Larochelle, I.~Lajoie, Y.~Bengio, and P.-A. Manzagol, ``Stacked
  denoising autoencoders: Learning useful representations in a deep network
  with a local denoising criterion,'' \emph{J. Mach. Learn. Res.}, vol.~11, pp.
  3371--3408, 2010.

\bibitem{Kingma2014AutoEncodingVB}
D.~P. Kingma and M.~Welling, ``Auto-encoding variational bayes,'' \emph{CoRR},
  vol. abs/1312.6114, 2014.

\bibitem{Xie2016UnsupervisedDE}
\BIBentryALTinterwordspacing
J.~Xie, R.~Girshick, and A.~Farhadi, ``Unsupervised deep embedding for
  clustering analysis,'' in \emph{Proceedings of The 33rd International
  Conference on Machine Learning}, ser. Proceedings of Machine Learning
  Research, vol.~48.\hskip 1em plus 0.5em minus 0.4em\relax PMLR, 20--22 Jun
  2016, pp. 478--487. [Online]. Available:
  \url{https://proceedings.mlr.press/v48/xieb16.html}
\BIBentrySTDinterwordspacing

\bibitem{Ji2019InvariantIC}
X.~Ji, A.~Vedaldi, and J.~F. Henriques, ``Invariant information clustering for
  unsupervised image classification and segmentation,'' \emph{2019 IEEE/CVF
  International Conference on Computer Vision (ICCV)}, pp. 9864--9873, 2019.

\bibitem{Huang2020DeepSC}
J.~Huang, S.~Gong, and X.~Zhu, ``Deep semantic clustering by partition
  confidence maximisation,'' \emph{2020 IEEE/CVF Conference on Computer Vision
  and Pattern Recognition (CVPR)}, pp. 8846--8855, 2020.

\bibitem{Wu2019DeepCC}
J.~Wu, K.~Long, F.~Wang, C.~Qian, C.~Li, Z.~Lin, and H.~Zha, ``Deep
  comprehensive correlation mining for image clustering,'' \emph{2019 IEEE/CVF
  International Conference on Computer Vision (ICCV)}, pp. 8149--8158, 2019.

\bibitem{Zhang2021SupportingCW}
D.~Zhang, F.~Nan, X.~Wei, S.~Li, H.~Zhu, K.~McKeown, R.~Nallapati, A.~O.
  Arnold, and B.~Xiang, ``Supporting clustering with contrastive learning,'' in
  \emph{NAACL}, 2021.

\bibitem{chen2020simple}
T.~Chen, S.~Kornblith, M.~Norouzi, and G.~Hinton, ``A simple framework for
  contrastive learning of visual representations,'' in \emph{International
  conference on machine learning}.\hskip 1em plus 0.5em minus 0.4em\relax PMLR,
  2020, pp. 1597--1607.

\bibitem{Krizhevsky2009LearningML}
A.~Krizhevsky, ``Learning multiple layers of features from tiny images,''
  University of Toronto, Toronto, Tech. Rep., 2009.

\bibitem{TaoTN21}
Y.~Tao, K.~Takagi, and K.~Nakata, ``Clustering-friendly representation learning
  via instance discrimination and feature decorrelation,'' in \emph{9th
  International Conference on Learning Representations (ICLR)}, 2021.

\bibitem{Chang2017DeepAI}
J.~Chang, L.~Wang, G.~Meng, S.~Xiang, and C.~Pan, ``Deep adaptive image
  clustering,'' \emph{2017 IEEE International Conference on Computer Vision
  (ICCV)}, pp. 5880--5888, 2017.

\bibitem{li2021sharper}
S.~Li and Y.~Liu, ``Sharper generalization bounds for clustering,'' in
  \emph{International Conference on Machine Learning}.\hskip 1em plus 0.5em
  minus 0.4em\relax PMLR, 2021, pp. 6392--6402.

\bibitem{pydi2019active}
\BIBentryALTinterwordspacing
M.~S. Pydi and V.~S. Lokhande, ``Active learning with importance sampling,''
  \emph{CoRR}, vol. abs/1910.04371, 2019. [Online]. Available:
  \url{http://arxiv.org/abs/1910.04371}
\BIBentrySTDinterwordspacing

\bibitem{Gao2006DiscoveringSO}
B.~J. Gao, O.~L. Griffith, M.~Ester, and S.~J.~M. Jones, ``Discovering
  significant opsm subspace clusters in massive gene expression data,'' in
  \emph{KDD '06}, 2006.

\bibitem{Cai2011HeterogeneousIF}
X.~Cai, F.~Nie, H.~Huang, and F.~Kamangar, ``Heterogeneous image feature
  integration via multi-modal spectral clustering,'' \emph{CVPR 2011}, pp.
  1977--1984, 2011.

\bibitem{Aggarwal2012ASO}
C.~C. Aggarwal and C.~Zhai, ``A survey of text clustering algorithms,'' in
  \emph{Mining Text Data}, 2012.

\bibitem{Basu2008ConstrainedCA}
S.~Basu, I.~Davidson, and K.~L. Wagstaff, \emph{Constrained Clustering:
  Advances in Algorithms, Theory, and Applications}.\hskip 1em plus 0.5em minus
  0.4em\relax Chapman and Hall/CRC, 2008.

\bibitem{Dasgupta2008HierarchicalSF}
S.~Dasgupta and D.~J. Hsu, ``Hierarchical sampling for active learning,'' in
  \emph{ICML '08}, 2008.

\bibitem{Wang2017ActiveLT}
M.~Wang, F.~Min, Z.~Zhang, and Y.~Wu, ``Active learning through density
  clustering,'' \emph{Expert Syst. Appl.}, vol.~85, pp. 305--317, 2017.

\bibitem{2013Spectral}
C.~Xiong, D.~Johnson, and J.~J. Corso, ``Spectral active clustering via
  purification of the $k$-nearest neighbor graph,'' in \emph{European
  Conference on Data Mining}, 2013.

\bibitem{Chien2019HS2AL}
E.~Chien, H.~Zhou, and P.~Li, ``Hs2: Active learning over hypergraphs with
  pointwise and pairwise queries,'' in \emph{AISTATS}, 2019.

\bibitem{Manduchi2021DeepCG}
L.~Manduchi, K.~Chin-Cheong, H.~Michel, S.~Wellmann, and J.~E. Vogt, ``Deep
  conditional gaussian mixture model for constrained clustering,'' in
  \emph{NeurIPS}, 2021.

\end{thebibliography}
% Generated by IEEEtran.bst, version: 1.14 (2015/08/26)

\end{document}